\pdfoutput=1
\documentclass[letterpaper]{article} 
\usepackage{aaai23}  
\usepackage{times}  
\usepackage{helvet}  
\usepackage{courier}  
\usepackage[hyphens]{url}  
\usepackage{graphicx} 
\usepackage{natbib}  
\usepackage{caption} 
\DeclareCaptionStyle{ruled}{labelfont=normalfont,labelsep=colon,strut=off} 
\usepackage{algorithm}
\usepackage{algorithmic}
\usepackage{subfig}

\usepackage{newfloat}
\usepackage{listings}
\floatstyle{ruled}
\newfloat{listing}{tb}{lst}{}
\floatname{listing}{Listing}
\title{A Large-Scale Comparative Study of Accurate COVID-19 Information versus Misinformation}
\author {
    Yida Mu,\textsuperscript{\rm 1}
    Ye Jiang, \textsuperscript{\rm 2}
    Freddy Heppell, \textsuperscript{\rm 1}
    Iknoor Singh, \textsuperscript{\rm 1}
    Carolina Scarton, \textsuperscript{\rm 1}
    Kalina Bontcheva, \textsuperscript{\rm 1}
    Xingyi Song \textsuperscript{\rm 1}
}
\affiliations {
    \textsuperscript{\rm 1} The University of Sheffield\\
    \textsuperscript{\rm 2} Qingdao University of Science and Technology\\
    {\{y.mu, frheppell1, i.singh,  c.scarton, k.bontcheva, x.song\}@sheffield.ac.uk} \\
    ye.jiang@qust.edu.cn 
}

\begin{document}

\maketitle
\begin{abstract}
The COVID-19 pandemic led to an infodemic where an overwhelming amount of COVID-19 related content was being disseminated at high velocity through social media. This made it challenging for citizens to differentiate between accurate and inaccurate information about COVID-19. This motivated us to carry out a comparative study of the characteristics of COVID-19 misinformation versus those of accurate COVID-19 information through a large-scale computational analysis of over 242 million tweets. The study makes comparisons alongside four key aspects: 1) the distribution of topics, 2) the live status of tweets, 3) language analysis and 4) the spreading power over time. An added contribution of this study is the creation of a COVID-19 misinformation classification dataset. Finally, we demonstrate that this new dataset helps improve misinformation classification by more than 9\% based on average F1 measure.
\end{abstract}


\section{Introduction}\label{sec1}
The COVID-19 pandemic was the first pandemic where social networks and mobile devices played a major role as sources of information in a rapidly evolving, dynamic context. Due to the large number of users and posts made daily on social platforms, a COVID-19 infodemic was created \cite{WHO2020a,WHO2020} where huge volumes of content (both accurate and mis/disinformation) were being published online daily. This made it challenging for citizens to find reliable information, as many fell victim to misinformation and turned to false treatments \citep{mehrpour2020toll, caceres2022impact,zhao2023prevalence} instead, or even started attacking medical workers \citep{orellana2023health, van2023attacks}.

In order to further our understanding of the infodemic, this paper undertakes a large-scale study of the statistical characteristics of accurate COVID-19 information compared to COVID-19 misinformation on Twitter. The paper aims to answer the following research questions:

\begin{itemize}
    \item Q1: What are the differences in topics and languages between accurate information and misinformation?
    \item Q2: What types of misinformation have social media platforms addressed?
    \item Q3: What is the spreading power of the different types of misinformation?
\end{itemize}

To help address these research questions, we developed an evidence-based COVID-19 misinformation classifier \cite{jiang2021categorising} based on a newly developed training set. Our experiments show that this new training data resulted in a significantly improved model that outperforms a state-of-the-art baseline model \cite{jiang2021categorising} by almost 0.2 in F1 measure score (0.51 vs 0.70) under leave-claim-out cross-validation\footnote{This evaluation method splits the training and testing sets based on the claim/topic, which is a realistic testing approach for evaluating the model's performance on unseen misinformation \cite{jiang2021categorising}}.

Next, we collected over 240 million COVID-19 related tweets and applied our best performing classifier to identify the misinformation posts automatically. This then enabled us to analyse the differences between misinformation and accurate posts. The statistical analysis makes comparisons alongside five dimensions: i) topical distribution, ii) spreading power, iii) deletion rate, iv) Bag-of-Words, and v) Linguistic Inquiry and Word Count Analysis.

\section{Related Work}\label{sec:relwork}

In this section we categorise COVID-19 misinformation statistical characteristic studies into three groups. The first category is the general COVID-19 information characteristic study, which investigates the characteristics of both non-mis- and mis-information but doesn't  split the results along these categories.  The second category is the COVID-19 misinformation characteristic study, where machine learning, external knowledge, or manual annotation is used to identify COVID-19 misinformation, and the statistical analysis is based only on misinformation. The third category is the COVID-19 non-mis- and mis-information statistical comparison study, which is the primary focus of this paper. This category investigates the statistical characteristics of both non-mis- and mis-information simultaneously, enabling a direct comparison between them.

\subsection{General COVID-19 information analysis}
\citet{singh2020first} conducted a study on COVID-19 related tweets from January 16 to March 15, 2020. The study reports the volume of tweets over various dimensions, including time, language, and geolocation. Additionally, the study investigates the statistics of word frequency, themes, popular myths, and URLs. Although Singh et al. did not focus specifically on misinformation tweets, the study reports statistics on tweets containing high-quality health sources and low-quality misinformation sources, based on NewsGuard\footnote{\url{https://www.newsguardtech.com/}}. Previous studies \cite{abdul2020mega,photiou2021social,iwendi2022covid,waheeb2022topic} have performed sentiment analysis \cite{medford2020infodemic,dashtian2021cml,chen2020eyes,lamsal2021design,gupta2021global,shi2020social}, hashtag analysis \cite{lamsal2021design,chen2020tracking,suarez2022use}, engagement \cite{cinelli2020covid}, and clustering analysis using topic modelling \cite{dashtian2021cml,gupta2021global,john2022covid} or pre-trained embeddings \cite{cinelli2020covid,biradar2022combating} on general COVID-19 related posts without specifically focusing on COVID-19 misinformation.

\subsection{COVID-19 misinformation analysis}
The statistical analyses on COVID-19 misinformation encompass sentiment \cite{sharma2020covid,cheng2021covid}, geolocation \cite{sharma2020covid}, topics modelling \cite{sharma2020covid,shi2020social}, sources \cite{sharma2020covid}, user analysis \cite{jain2021state}, engagement level \cite{jain2021state}, and hashtag and word frequency \cite{sharma2020covid,dharawat2020drink,jain2021state,cheng2021covid} have also been conducted in previous studies. In addition, \citet{brennen2020types} conducted a misinformation topic analysis that can be applied to prioritise the fact-check resources to debunk the most harmful kind of misinformation topics. Brennen et al. defined nine different types of misinformation topics and found that the most spread misinformation is related to public authority actions in the early stages of the COVID-19 pandemic.  \citet{song2021classification} extend this study on a larger scale, and report the topic evolution through different time periods. In this paper, we continue the topic study \cite{brennen2020types,song2021classification} but in different dimensions which include its spread, live tweet status and sentiment analysis.

Next, we discuss some of the approaches used for identifying misinformation from abundant COVID-19 related information. Some of the previous methods include, 1) Manual annotation \cite{jiang2021categorising,brennen2020types}: the most reliable but also the most expensive approach. Therefore, analysis based on this approach is often limited to a small scale. 2) Credibility based classification \cite{sharma2020covid,cheng2021covid}: which identifies misinformation based on the credibility of the author and/or the source (e.g. shared URLs). Compared to manual annotation, the credibility approach is a much cheaper alternative to adapt misinformation analysis to a larger scale; however, the major drawback of this approach is identifying misinformation indirectly. \citet{jiang2021categorising} notice that misinformation is often shared from highly credible sources. 
3) Style based machine learning classification \cite{cheng2021covid,abdelminaam2021coaid,kar2020no}: style based classifiers identify misinformation based on the writing styles of the text. This approach is effective and often performs well with the misinformation (theme/topics) already covered in the training data.  On the other hand, the style-based classifier is less effective in identifying misinformation that is not covered in the training data \cite{jiang2021categorising}. 4) Evidence-based machine learning classification \cite{jiang2021categorising,hossain-etal-2020-covidlies,pan2018content,hu2021compare}: in comparison with the style based classification solely relying on the input text, evidence-based classification also requires external knowledge as evidence to aid the misinformation classification. The external knowledge could be a professional debunked misinformation database \cite{jiang2021categorising,hossain-etal-2020-covidlies}  or knowledge graph built from true \cite{pan2018content,hu2021compare} or misinformation \cite{pan2018content}. This approach often provides more reliable misinformation detection results and the evidence is also provided with the results. In this paper, we also apply an evidence-based classification approach for misinformation identification.

\subsection{Misinformation Comparision Analysis}

\citet{cui2020coaid} analysed sentiment and hashtag frequency of mis- and non-mis- information and find that COVID-19 misinformation leans more towards negative sentiment compared to non-misinformation, and also hashtag distributions are significantly different. \citet{memon2020characterizing} studied the Twitter user’s network density, bot ratio and sociolinguistics in the post and demonstrate that misinformed users have a higher bot ratio and a denser network (i.e. accordance opinions are more easily accepted and it is also more difficult to have opposing opinions accepted in a denser network). The study also finds some sociolinguistic differences between mis and non-mis- information, but the results are not significant and are inconclusive. \citet{micallef2020role} conducted a case study to investigate the spread of misinformation and debunks related to some pre-defined COVID-19 topics (e.g.,  5G conspiracy theories), which may lead to hesitation towards COVID-19 vaccination \citep{mu2023vaxxhesitancy}. \citet{burel2020co} did a post-hoc causality analysis to study the spread of COVID-19 misinformation and fact-checks on Twitter, while \citet{recuero2022bolsonaro} used Crowdtangle to study the spread of COVID-19 disinformation and fact-checks on Facebook. \citet{silva2020predicting} and \citet{kouzy2020coronavirus} both study the engagement level between mis and non-mis- information, however, the conclusions from these two studies are different. \citet{kouzy2020coronavirus} observed that there is no difference in the engagement level between the two groups, but \citet{silva2020predicting} notice that misinformation often has a higher engagement level. The difference is possibly due to the different data collection methods.

\section{Data and Misinformation Classification}
\label{sec2}

The data in this work is collected from Twitter via Twitter Stream API\footnote{ \url{https://developer.twitter.com/en/docs/twitter-api}} using keywords associated with COVID-19 (e.g. covid and covid-19, the complete list is in Table \ref{tab:keyword}). The data was collected from March 2020 to July 2021, and we collected a total of 242,823,999 tweets. However, we only considered the source tweets for analysis and therefore excluded retweets and replies, resulting in 6,691,181 remaining source tweets that were used in this study.

Our approach for detecting misinformation is based on evidence-based classification, adapted from \citet{jiang2021categorising} \footnote{We used a coarse-grained BERT pairwise classification model}. The classifier is trained through a pairwise approach that similar to BERT next sentence prediction \citet{devlin2019bert}. The input pairs consist: 1) \textbf{Claim:} a verified misinformation claim from a professional fact-checker and 2) \textbf{Tweet:} the text of the tweet to be classified. Please note the classification process solely relies on the tweet text for categorisation, and does \textbf{not} utilise any additional contextual information, such as quoted text, images, or external web pages referenced within the tweet. The classifier predicts whether the tweet text belongs to one of three classes: a) \textbf{Misinformation}, b) \textbf{Debunk}, or c) \textbf{Irrelevant}. We have enhanced the original model by adding two additional steps: 1) Data enrichment and 2) Results filtering to improve classification accuracy. We describe these steps in detail in Section `Data Enrichment' and Section `Misinformation Detection'.

\subsection{Data Enrichment} 
\label{sec:dataenrich}

In addition to the training data outlined in \citet{jiang2021categorising}, we expanded our dataset by collecting training data from three additional sources: 1) the CovidLies \cite{hossain2020covidlies} dataset, 2) our own collection of tweets with external knowledge, and 3) the IFCN Poynter website. The process of enriching the dataset with data from these sources will be discussed in this section.

The {\bf CovidLies} dataset provides a manually annotated evidence-based dataset for misinformation classification. We converted its label set directly into the labels used in \citet{jiang2021categorising}: {\it Misinformation}, {\it Debunk}, and {\it Irrelevant}. However, many tweets were deleted by the time we reconstructed the CovidLies data using Twitter API, leaving us with only 87 {\it Misinformation}, 87 {\it Debunk}, and 3,142 {\it Irrelevant} tweets.

We employed external knowledge to label our {\bf in-house tweet collection} to create a `silver standard' training dataset. First, select all the source tweets (i.e. excluding replies or retweets) and generate candidate labels for those tweets  using the original \cite{jiang2021categorising} trained classifier. For the next step, we only keep {\it Misinformation} and {\it Debunk} tweets based on the output label form classifier with a confidence score (i.e. model's softmax output) greater than 0.7. The threshold was determined by manually examining a subset of samples.

In the third step, we used Twitter's COVID-19 misleading information policy to identify deleted tweets likely to be {\it Misinformation}. Any candidate {\it Misinformation} tweet that had been removed from Twitter was selected as a final {\it Misinformation} sample.

For {\it Debunk} samples, we noticed that debunking tweets often reference articles published on professional fact-checker websites. We utilised the IFCN Poynter debunking data as a knowledge source to select any remaining {\it Debunk} candidate tweets containing URLs linked to the IFCN Poynter debunks as enriched {\it Debunk} training samples.

Finally, we selected {\it Irrelevant} training samples to correct false positives caused by the classifier. These samples were extracted from the {\it Misinformation} candidates but posted from highly credible accounts, such as WHO. The full list of credible accounts is provided in Table \ref{tb:credaccot}.

\begin{figure*}[!t]
\centering
\includegraphics[scale=0.4]{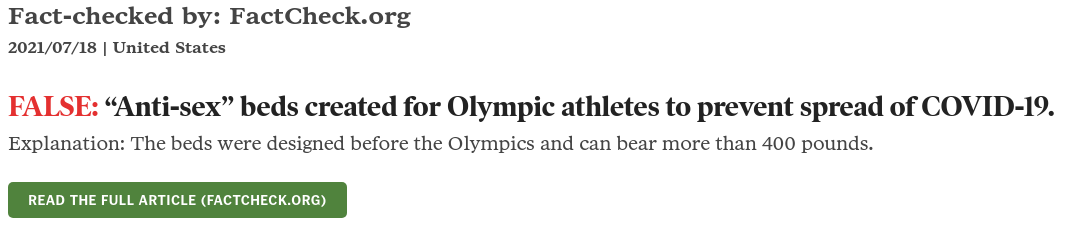}

\label{fig:IFCNweb}
\end{figure*}

\begin{figure*}[!t]
\centering
\includegraphics[scale=0.4]{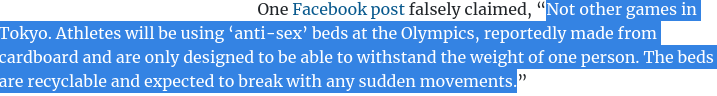}
\vspace{-2mm}
\caption{{\bf (a)}: Screenshot of an IFCN debunk post. The post includes 1) fact checking organisation, 2) misinformation claim, 3) explanation of why the claim is false and 4) the link to full debunk article. {\bf(b)}: Partial screenshot of full debunk article of the IFCN debunk post.}
\vspace{-4mm}
\label{fig:IFCNori}
\end{figure*}

The {\bf IFCN Poynter website} provides a valuable source for enriching our training samples. As shown in Figure \ref{fig:IFCNori}, each debunk on the website includes a misinformation claim, an explanation of why it is false, and a link to the full debunk article. For example:

To extract training samples from this source, we collect all the explanations and use them as {\it Debunk} samples. For {\it Misinformation} samples, we need to access the full article of the debunk, as shown in Figure \ref{fig:IFCNori}. The original source misinformation claims are often referenced in the article (highlighted in blue in Figure \ref{fig:IFCNori}). We use regular expressions to extract this text and label it as {\it Misinformation} in our enriched dataset.

\begin{table}[!t]
\small
\centering
\begin{tabular}{|l|c|c|c|}
\hline 
\bf DataSet & \bf IRRELEVANT &\bf DEBUNK &\bf MISINFO \\
\hline 
Jiang et al.    & 1066 & 194 & 522 \\
CovidLies               & 3142 & 87 & 123 \\
Tweet Collect       & 5757 & 3692 & 670 \\
IFCN                    & 0 & 1892 & 96 \\
Total                   & 9965 & 5865 & 1411 \\
\hline 
\end{tabular}
\vspace{-2mm}
\caption{The statistic of enriched COVID-19 misinformation classification training dataset.}
\vspace{-4mm}
\label{tb:trainingsta}
\end{table}

\subsubsection{Enrichment Data Cleaning} \label{sec:dataclean}

Since the classifier used in this study is trained on pairwise input (claim and tweet text pair), tweet text can only be accurately labelled as {\it Misinformation} if it matches the paired claim. However, in the data enrichment process using in-house and IFCN data, there is no guarantee that the pairing process will be accurate. To improve the quality of the enriched data and reduce the number of mismatches between claim and text pairs, a pair cleaning process is conducted. Specifically, the Stanford OpenIE SVO parser \cite{angeli2015leveraging} is used to extract the Subject and Object in both the claim and the tweet text. Samples are discarded if neither the claim Subject nor the claim Object is mentioned in the tweet text. Table \ref{tb:trainingsta} shows the statistics of training samples after the cleaning process.

\subsubsection{Model Training and Performance} \label{sec:modelresults}

The training process of the model is based on \citet{jiang2021categorising}, and we will briefly explain the key settings in this section. For more detailed training settings, please refer to the paper.

The classification model is a fine tuned COVID-Twitter Pre-Trained \cite{muller2020covid} BERT Large \cite{devlin2019bert}. It contains 24 transformer layers, and in our experiment, only the parameters in the last transformer encoding layer are unlocked for fine-tuning, while the remaining BERT weights remain frozen. Similar to the BERT next sentence prediction task, the input to the model is constructed as `[CLS] + Claim + [SEP] + Tweet\_Text + [SEP]', and the Softmax classifier predicts the probability of labels based on the pairwise [CLS] representation.

We conduct five-folds `Leave-Claim-Out Cross Validation' described in \citet{jiang2021categorising} to compare enrichment performance. Note that the five-folds are only conducted on the original \citet{jiang2021categorising} annotated dataset. The enriched data is only added as supplemental training data, therefore, the total number of test samples is the same as in the experiment described in  \citet{jiang2021categorising}. The purpose of Leave-Claim-Out Cross Validation is to test the classification performance for unseen misinformation. To ensure a fair comparison with the non-enriched model, we manually removed claims from the enriched dataset that were already present in the original annotated dataset in \citet{jiang2021categorising}, so that the test claims were unseen during training.

\begin{table}[!t]
\centering
\begin{tabular}{|l|c|c|}
\hline
 & \textbf{Non-enriched}  & \textbf{Enriched} \\
\hline
Accuracy       & 0.64 & 0.70 \\
Avg. F1        & 0.57 & 0.66 \\
Debunk F1      & 0.47 & 0.56 \\
IRRELEVANT F1  & 0.72 & 0.72 \\
Misinformation F1     & 0.51 & 0.70 \\
\hline
\end{tabular}
\vspace{-2mm}
\caption{The macro average of five fold Leave-Claim-Out Cross Validation results. The number of Non-enriched Classifier is directly borrowed from \citet{jiang2021categorising}.}  
\vspace{-4mm}
\label{tb:norefres}
\end{table}

Compared to the non-enriched classifier, our enriched classifier yields an overall F1 measure of 0.66, which is more than 9\% higher than the non-enriched classifier's overall F1 measure of 0.57. Furthermore, the F1 measure of the {\it Misinformation} class is significantly increased from 0.51 to 0.70 by our enriched classifier.

\subsection{Misinformation Detection} \label{sec:resfil}

In the misinformation detection process, we used 12,748 claims collected from the IFCN Poynter website as input for our pairwise classifier. The claim collection process followed the same procedure as in previous studies \cite{song2021classification, jiang2021categorising}.

However, since our classifier requires pairwise inputs, classifying the 6 million source tweets against the 12,748 IFCN claims would result in 72 billion pairs, which is computationally expensive. To reduce the computational cost and speed up the classification process, we conducted an additional candidate pair selection step. Here's an overview of our misinformation extraction procedure:

\begin{enumerate}
    \item We index the 6 million source tweets using Elastic Search.
    \item The 12,748 IFCN claims are used as queries to search for relevant tweets in the index. We use the BM25 ranking algorithm \cite{robertson1995okapi} to rank the tweets and select the top 20,000 tweets for each query based on their BM25 score.
    \item After retrieving the top 20,000 tweets using BM25, we conduct reranking based on the cosine similarity between tinyBERT embeddings of the queries and tweets. This allows us to select the top 1,000 tweets that are semantically similar to the IFCN claim, which serves as candidate pairs for the classification process.
    \item Misinformation may also be time-sensitive. The claim may become true or false in different time periods. For example, in early March 2020, there was misinformation claiming that Trump was infected with COVID-19\footnote{\url{https://www.poynter.org/?ifcn_misinformation=trump-faints-is-infected-with-coronavirus}}, which remained misinformation until October 2020 when the White House COVID-19 outbreak meant Donald Trump actually tested positive for COVID-19\footnote{\url{https://www.forbes.com/sites/elanagross/2020/10/02/white-house-outbreak-here-are-the-people-who-have-tested-positive-for-covid-near-president-trump/?sh=43ab1d76799a}}. Therefore, we only consider tweets that were posted within a date range of ten weeks before and two weeks after the IFCN claim debunk date. Candidate tweets that fall outside of this period will be filtered out.
    \item The misinformation classifier (described in Section `Model Training and Performance') is applied to the remaining candidates.  A post-process step is conducted to ensure the precision of the misinformation extraction. The final set of misinformation tweets is selected if the classifier predicted label is `misinformation' and satisfies one of the following conditions: 1) the classifier confidence score is greater than, or equal to 0.95, or 2) the tweet text shares the same subject or object with the paired IFCN claim. The subject and object are extracted using the Stanford OpenIE SVO parser. In total, we collected 14,058 misinformation tweets.

\end{enumerate}

\section{Misinformation Analysis}\label{sec:misinfoanalysis}
This section presents a statistical analysis of 14,058 source misinformation tweets, including their topical distribution, emotional categories, and distributional statistics. For comparison, we also conduct the same analysis for non-misinformation tweets, which consist of 20,000 randomly selected source tweets that were not classified as `misinformation' by the classifier.

\subsection{Topical Distribution}
\label{sec:topicalDistribution}

\begin{figure}[!t]
\centering
\includegraphics[scale=0.4]{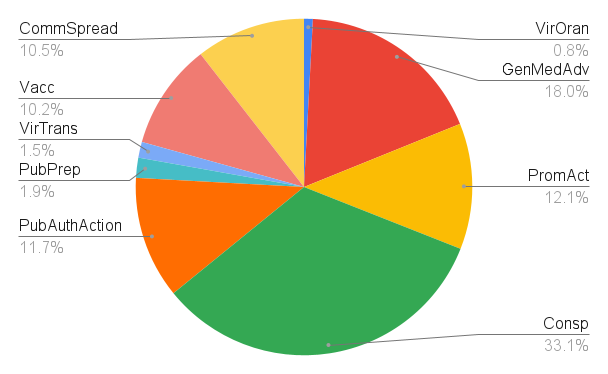}
\vspace{-2mm}
\caption{Misinformation tweets topic distribution}
\vspace{-4mm}
\label{fig:misinfopie}
\end{figure}

\begin{figure}[!t]
\centering
\includegraphics[scale=0.4]{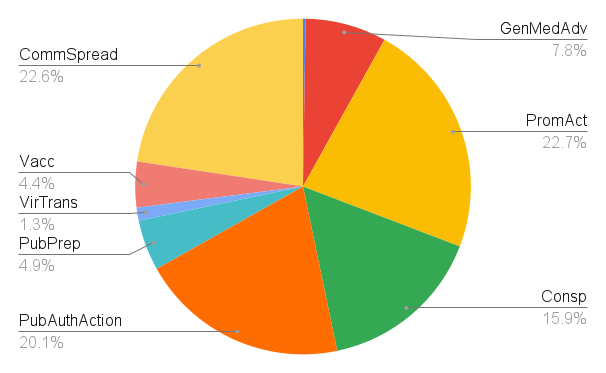}
\vspace{-2mm}
\caption{Non-misinformation tweets topic distribution}
\vspace{-4mm}
\label{fig:nonmisinfopie}
\end{figure}

\begin{figure}[!t]
\centering
\includegraphics[scale=0.4]{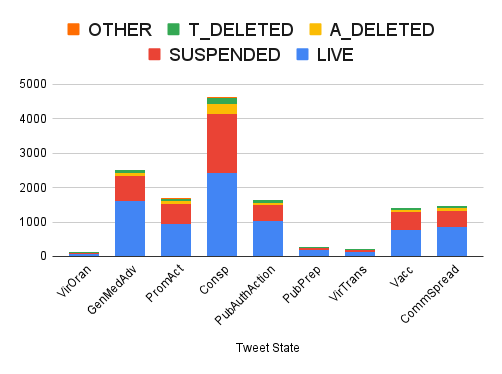}
\vspace{-2mm}
\caption{Misinformation tweets topic distribution}
\vspace{-4mm}
\label{fig:misinfobar}
\end{figure}

\begin{figure}[!t]
\centering
\includegraphics[scale=0.4]{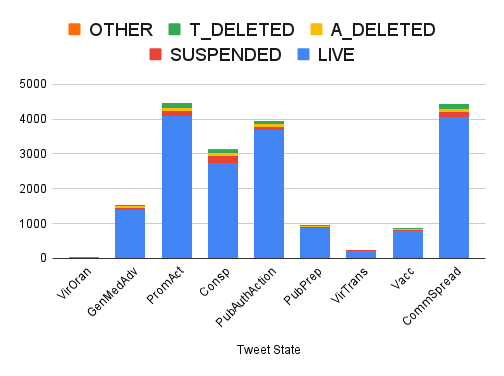}
\vspace{-2mm}
\caption{Non-misinformation tweets live state}
\vspace{-4mm}
\label{fig:nonmisinfobar}
\end{figure}

\begin{figure*}[!t]
\centering
\includegraphics[scale=0.5]{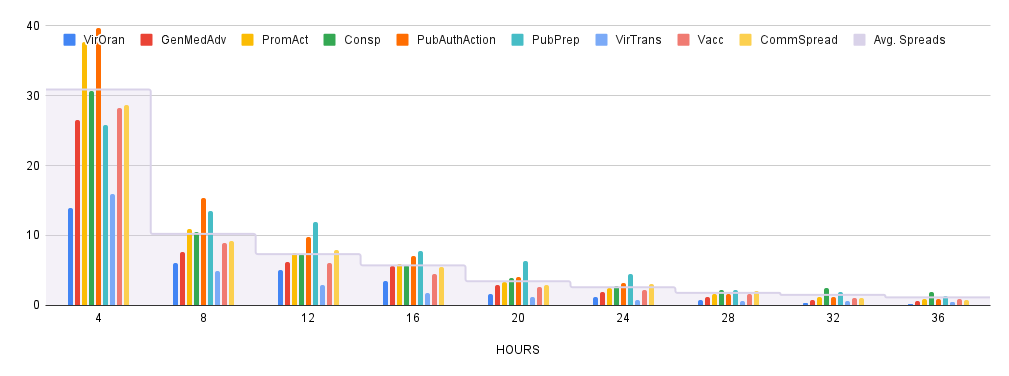}
\vspace{-2mm}
\caption{Spreading power of misinformation tweets.}
\vspace{-4mm}
\label{fig:misinfolast}
\end{figure*}

\begin{figure*}[!t]
\centering
\includegraphics[scale=0.5]{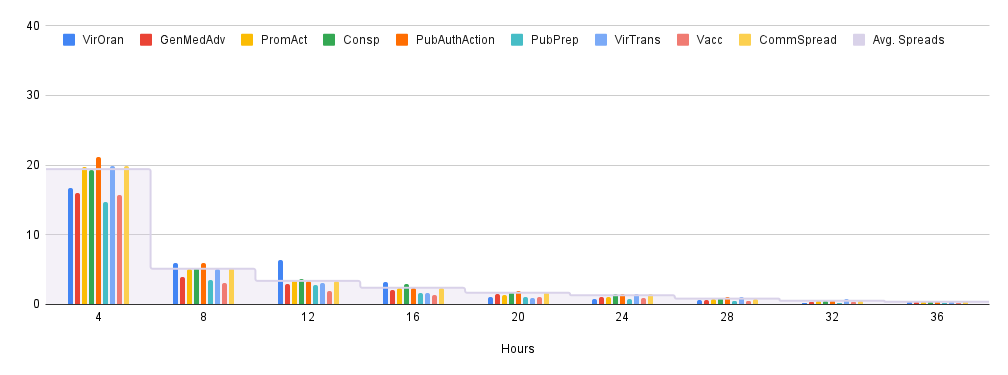}
\vspace{-4mm}
\caption{Spreading power of non-misinformation tweets. Average number of spreads of source misinformation and non-misinformation tweets (retweets or quotes) in 4 hour intervals. The coloured columns are the average number of spreads in each topic, and the purple stepped line shows the average number of spreads including all topics.}
\vspace{-4mm}
\label{fig:nonmisinfolast}
\end{figure*}

Figure \ref{fig:misinfopie} and Figure \ref{fig:nonmisinfopie} are the topic distribution of source misinformation and non-misinformation tweets. 

The topic distribution of source misinformation tweets is notably different from that of non-misinformation tweets. The most frequently mentioned topic in the misinformation tweets is `conspiracy theory’, accounting for almost one third (33.1\%) of all the misinformation tweets. The second-largest proportion of the misinformation tweets pertains to `general medical advice' (18\%). Conversely, the most commonly mentioned topics in non-misinformation tweets are `prominent actors' (22.7\%), `community spread and impact' (22.6\%) and `public authority action' (20.1\%). 

Figure \ref{fig:misinfobar} and Figure \ref{fig:nonmisinfobar} display bar charts that show the number of tweets in each topic, with colours indicating the live status of the tweet as of October 28, 2021. The live status of the tweets was determined by revisiting them via the Twitter API. The colours used in the charts represent the following live statuses: blue indicates that the tweet was still available on the day of the revisit, red indicates that the account used to post the tweet has been suspended by Twitter, Account deleted indicates that the account used to post the tweet has been deleted for an unknown reason, `Tweet deleted' indicates that the tweet has been deleted for an unknown reason, and `Other' indicates that the tweet was not accessible on the day of the revisit due to privacy settings.

The live statuses of misinformation and non-misinformation tweets exhibit significant differences. More than 40\% of the misinformation tweets were not accessible during revisit, and the primary reason was account suspension, with a suspension rate of 33.1\%. In contrast, only 8.8\% of non-misinformation tweets were inaccessible, and the account suspension rate was much lower at 3.7\%.

The spread of conspiracy theory misinformation seems to have garnered significant attention from Twitter, with approximately half of the conspiracy theory misinformation tweets being removed from the platform. In particular, the account suspension rate for tweets about conspiracy theories is 37\%. On the other hand, misinformation tweets about the `virus origin' seem to have received the least attention, with almost 70\% of such tweets still being available on the platform and the account suspension rate being only 23.7\%.

\begin{figure}[!t]
\centering
\includegraphics[scale=0.3]{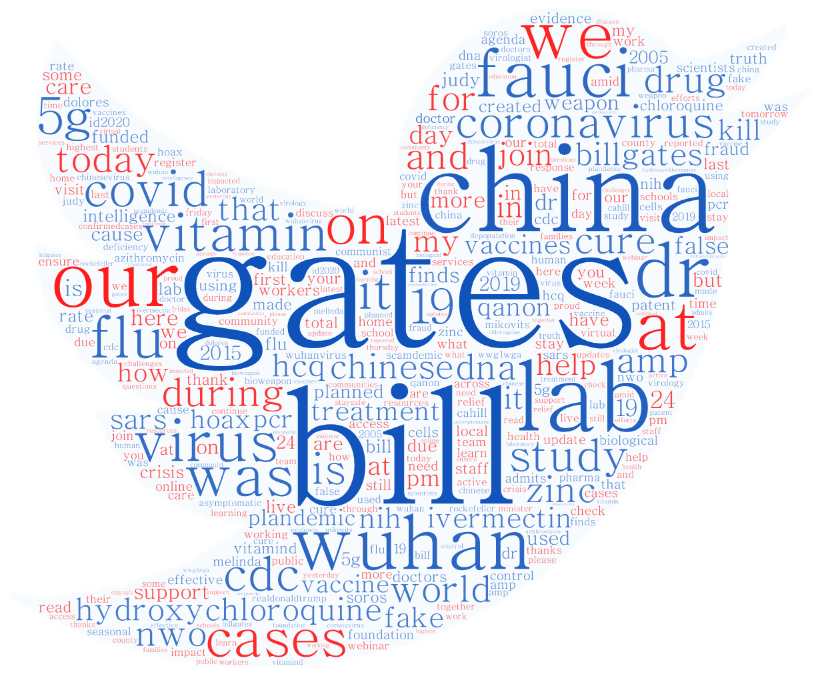}
\vspace{-2mm}
\caption{Bag-of-Words features associated with Non-misinformation (Red) and Misinformation (Blue). Note that bigger font size denotes higher Pearson's Correlation R.}
\label{fig:word cloud}
\end{figure}

\begin{table}[!t]
\centering
\scriptsize
\begin{tabular}{|l|l||l|l|}
\hline
\multicolumn{4}{|c|}{{\bf Bag-Of-Words}} \\ \hline
{\bf Non-misinformation}            & {\bf r}         & {\bf Misinformation}           & {\bf r}       \\ \hline
Our        & 0.138 & Gates      & 0.204 \\ \hline
Cases      & 0.127 & Bill       & 0.177 \\ \hline
During     & 0.125 & China      & 0.172 \\ \hline
Today      & 0.121 & hydroxychloroquine  & 0.171 \\ \hline
We         & 0.117 & Wuhan       & 0.164 \\ \hline
At         & 0.107 & Coronavirus & 0.159 \\ \hline
On         & 0.101 & Lab         & 0.149 \\ \hline
And        & 0.098 & Fauci      & 0.144 \\ \hline
Support    & 0.098 & Vitamin    & 0.143 \\ \hline
Help       & 0.093 & Virus      & 0.143 \\ \hline            

\end{tabular}
\caption{N-grams associated with Non-mis- and Mis- information sorted by Pearson's correlation ($r$) between the TF-IDF frequency and the labels ($p < .05$).}
\vspace{-4mm}
\label{t:bow}
\end{table}

\begin{table}[!t]
\centering
\scriptsize
\begin{tabular}{|l|l||l|l|}
\hline
\multicolumn{4}{|c|}{{\bf LIWC}} \\ \hline
{\bf Non-misinformation}        & {\bf r}            & {\bf Misinformation}         & {\bf r}          \\ \hline
Affiliation         & 0.200        & Health             & 0.164      \\ \hline
Time                & 0.186        & Biological Processes & 0.142      \\ \hline
Emotional Tone      & 0.170        & Negative Emotion   & 0.131      \\ \hline
Relativity          & 0.170        & Anger              & 0.126      \\ \hline
Prepositions        & 0.148        & Quotation Marks    & 0.123      \\ \hline
Authentic           & 0.136        & Past Focus         & 0.113      \\ \hline
Drives              & 0.133        & Death              & 0.109      \\ \hline
Positive Emotion    & 0.126        & Semicolons         & 0.105      \\ \hline
We                  & 0.126        & Causation          & 0.093      \\ \hline
Social              & 0.124        & Dashes             & 0.076      \\ \hline
\end{tabular}
\caption{LIWC categories associated with Non-mis- and Mis- information sorted by Pearson's correlation ($r$) between the normalised frequency and the labels ($p < .05$).}
\label{t:liwc}
\end{table}

Figure \ref{fig:misinfolast} and Figure \ref{fig:nonmisinfolast} show the average number of spreads (i.e. retweets or quotes) of source misinformation and non-misinformation tweets in 4-hour intervals. 
Overall, the analysis reveals that misinformation spreads more quickly than non-misinformation. Specifically, the average spread power of a misinformation tweet during the first 36 hours is 64.5, which means that the tweet is retweeted or quoted an average of 64.5 times in that period. By contrast, non-misinformation tweets have a much lower average spread power of 34.8 during the first 36 hours.

During the first 4 hours, the spread power of misinformation tweets is highest, with a value of 30.8. The topics of `prominent actor' and `public authority action' have notably higher spread power than other topics, with values of 37.6 and 39.6 spreads, respectively. In contrast, the spread of non-misinformation is more evenly distributed across different topics, and the average spread power in the first 4 hours is 19.4.

After 4 hours, the number of spreads reduces significantly. In the next 4-hour period, the average spread power of misinformation drops to 10.2, and that of non-misinformation drops to 5.09. During this period, `public authority action' related misinformation still receives significantly higher spreads (15.3 spreads) than other topics. In the period between 12 and 28 hours, `public preparation' misinformation remains a hot topic, where an average tweet on this topic has 32.8 spreads. Moreover, `conspiracy theory' has a longer spreading period than other topics, as it is the only topic that still has more than two spreads even after 32 hours.

\subsection{Linguistic Analysis} \label{sec:liwcanalysis}
In this section, we compare linguistic features between misinformation and non-misinformation tweets. For that purpose, we follow a similar approach to previous studies by \citet{Schwartz2013} and extract the top 10 correlated features using Bag-Of-Words and Linguistic Inquiry and Word Count (LIWC) \cite{pennebaker2015development}. 
We conduct the univariate Pearson's Correlation test to determine which linguistic features, such as individual BOW tokens and LIWC categories, are highly correlated with categorical variables (i.e., misinformation or non-misinformation), in line with recent research in computational misinformation analysis \citep{mu2020identifying,mu2022identifying}. The reported linguistic features are statistically significant and show correlations with both misinformation and non-misinformation categories (with a p-value of less than 0.05).

\paragraph{Bag-Of-Words}
We established specific criteria to select the Bag-Of-Words (BOW) features, which are as follows: i) the word's document frequency must be between three and 40\% of the total tweets (i.e. $3 < df < 0.4*t$ where $df$ is the document frequency, $t=32,748$ is the total number of tweets used for comparison study). ii) The words are then scored based on the TF-IDF score, and we keep the top 5,000 words as the BOW features for this analysis.

The BOW analysis results are presented in Table~\ref{t:bow}, where Pearson’s correlations are reported in column {\bf r}. The top 10 BOWs for misinformation tweets align with the findings from the topical distribution analysis discussed in Section `Topical Distribution'. The top 10 BOWs are mainly associated with `conspiracy theories' (e.g. \emph{Gates}, \emph{Bill}\footnote{\url{https://www.politifact.com/factchecks/2020/may/20/facebook-posts/no-gates-foundation-isnt-pushing-microchips-all-me/}}, \emph{Lab}, \emph{China} \footnote{\url{https://www.poynter.org/?ifcn_misinformation=coronavirus-has-been-originated-in-a-laboratory-linked-to-chinas-biowarfare-program-2}}) and `general medical advice' (e.g. \emph{hydroxychloroquine}\footnote{\url{https://www.poynter.org/?ifcn_misinformation=hydroxychloroquine-completely-cures-people-infected-with-covid-19}}, \emph{Vitamin}). Compared to non-misinformation, the top 10 BOWs all related to common words (e.g. \emph{Our, We, On, And}).

\begin{table*}[]
\begin{tabular}{|l|l|l|l|l|}
\hline
\#blackfungus         & \#covid19vaccination & \#socialdistancing       & \#chinesebioterrorism & \#greatreset        \\ \hline
\#chinesevirus     & \#covid19vaccine     & \#stayathome             & \#chinesevirus       & \#hydroxichloroquine  \\ \hline
\#corona           & \#covid2019          & \#stayhomestaysafe       & \#ciavirus           & \#hydroxychloroquine  \\ \hline
\#corona19         & \#covid2019uk        & \#staysafe               & \#coronabollocks     & \#idonotconsent       \\ \hline
\#corona2019       & \#covid\_19          & \#unite2fightcorona      & \#coronacon          & \#israelvirus         \\ \hline
\#coronapandemic   & \#covid\_19\_uk      & \#wearamask              & \#coronafacts        & \#kungflu             \\ \hline
\#coronasecondwave    & \#covid\_2019\_uk    & \#workfromhome           & \#coronafakenews      & \#mildsymptoms      \\ \hline
\#coronaupdate     & \#covidemergency     & \#wuhanvirus             & \#coronafraud        & \#nwo                 \\ \hline
\#coronav          & \#covidemergency2021 & \#5g                     & \#coronahoax         & \#nwoevilelites       \\ \hline
\#coronavaccine    & \#covidhelp          & \#5gcoronavirus          & \#coronasymptoms     & \#nwoevilplans        \\ \hline
\#coronavirus      & \#covidresources     & \#americavirus           & \#coronavillains     & \#nwovirus            \\ \hline
\#coronavirus19    & \#covidsecondwave    & \#astrazeneca            & \#coronavirus5g      & \#obamagate           \\ \hline
\#coronavirus2019  & \#covidsos           & \#ccpvirus               & \#coronaviruscoverup & \#oxfordvaccine       \\ \hline
\#coronavirusoutbreak & \#coviduk            & \#chinaliedandpeopledied & \#coronavirusfacts    & \#plandemic         \\ \hline
\#coronaviruspandemic & \#covidvaccination   & \#chinaliedpeopledied    & \#covid19symptoms     & \#preventnwo        \\ \hline
\#coronavirustruth & \#covidvaccine       & \#chinaliespeopledied    & \#covidiots          & \#remdesivir          \\ \hline
\#coronavirusupdates  & \#covid?19           & \#chinesebioterrorism    & \#covidsymptoms       & \#reopenbritain     \\ \hline
\#covid            & \#lockdown           & \#chinesevirus           & \#cronyvirus         & \#resistthegreatreset \\ \hline
\#covid-19         & \#lockdown2021       & \#ciavirus               & \#deepstatevirus     & \#scamdemic           \\ \hline
\#covid-19-uk      & \#pandemic           & \#chinaliedandpeopledied & \#depopulation       & \#sorosvirus          \\ \hline
\#covid19             & \#remotework         & \#chinaliedpeopledied    & \#endthelockdown      & \#wholiedpeopledied \\ \hline
\#covid19uk        & \#sarscov2           & \#chinaliespeopledied    & \#endthelockdownuk   & \#wuhanvirus          \\ \hline
\end{tabular}%
\caption{The full list keywords used for Twitter collection though Twitter API}
\vspace{-4mm}
\label{tab:keyword}
\end{table*}

\begin{table}[!htbp]
\small
\centering
\begin{tabular}{|l|c|}
\hline
   \textbf{Account Name} & \textbf{Display Name} \\
\hline
ANI & ANI   \\
GT & Global Times \\
roinnslainte &  Department of Health  \\
WHO & World Health Organization (WHO)   \\
wef & World Economic Forum   \\
GHS & Global Health Strategies   \\
AFP & AFP News Agency   \\
UN & United Nations   \\
EMRO & WHO Eastern Mediterranean Regional Office   \\
\hline
\end{tabular}
\caption{The full list of credible accounts applied for the {\it IRRELEVANT} training sample enrichment.}
\vspace{-4mm}
\label{tb:credaccot} 
\end{table}

\paragraph{LIWC}
The Linguistic Inquiry and Word Count (LIWC) \cite{pennebaker2015development} is a dictionary of around 6,400 words that are mapped to 93 manually created lexical categories. LIWC has been widely used in computational social science. For this study, we used version 2015 of LIWC \cite{pennebaker2015development}. Examples of words for each category can be found in the user manual\footnote{\url{https://repositories.lib.utexas.edu/bitstream/handle/2152/31333/LIWC2015_LanguageManual.pdf}}. The top 10 LIWC categories, sorted by Pearson's Correlation ({\bf r}) between the normalised frequency and the labels (i.e., Non-mis vs. Misinformation), are shown in Table~\ref{t:liwc}. We observed that LIWC categories such as \emph{Health}, \emph{Biological Processes}, and \emph{Death} are highly correlated with misinformation.

Furthermore, Table~\ref{t:liwc} displays the top 10 LIWC categories sorted by Pearson's Correlation ({\bf r}) between the normalised frequency and the tweet categories (i.e., non-misinformation vs. misinformation). Misinformation is highly correlated with categories such as `Anger', `Negative Emotion', and `Death'. In contrast, non-misinformation is more correlated with opposite categories, such as `Positive Emotion', `Authentic', and social categories (`Social', `We', `Affiliation').

\section{Conclusion}\label{sec13}
This paper presents a comprehensive comparison study between non-mis- and mis- COVID-19 information, using a machine learned classifier to extract misinformation. To ensure the accuracy of misinformation classification, we introduced a data enrichment process and post-process steps.

The results of our comparison study show that there are clear differences between COVID-19 misinformation and non-misinformation tweets. Specifically, we found that non-misinformation tweets predominantly focus on prominent actors and community spread, whereas approximately one-third of misinformation source tweets are related to conspiracy theory. 

Our linguistic analysis supports the findings from the topical analysis. The top 10 Bag-Of-Words features associated with misinformation are related to conspiracy theory. Moreover, our LIWC analysis indicates that misinformation is frequently associated with `negative emotion', `anger', and `death', while non-misinformation is associated with `positive emotion', `authenticity', and `social' categories.

Addressing misinformation related to conspiracy theory appears to be the primary concern of social media platforms, as evidenced by the removal of over 40\% of related misinformation tweets on Twitter. Conversely, tweets related to the topic of virus origin have received the least attention, with nearly 70\% of such tweets still accessible.

Misinformation tweets have a spreading power 158\% higher than non-misinformation tweets. Notably, tweets related to conspiracy theory have a longer spreading period than other topics, with more than two spreads even after 32 hours.

\section{Acknowledgements}
This work has been co-funded by the European Union under the Horizon Europe vera.ai (grant 101070093) and Vigilant (grant 101073921) projects and the UK’s innovation agency (Innovate UK) grants 10039055 and 10039039. We would like to thank all the anonymous reviewers for their valuable feedback.
\bibliography{covidanalysis}
\end{document}